\documentclass[runningheads]{llncs}
\usepackage[T1]{fontenc}

\usepackage{amsmath} 
\usepackage{amssymb}  
\usepackage{multirow}
\usepackage{tabularx}
\usepackage{booktabs}
\usepackage{subcaption}
\usepackage{graphicx}
\usepackage{arydshln}   
\usepackage{gensymb}    
\usepackage{dashrule}   

\usepackage{hyperref}
\hypersetup{
    colorlinks=true,
    linkcolor=blue,
    filecolor=magenta,      
    urlcolor=cyan,
    pdftitle={Overleaf Example},
    pdfpagemode=FullScreen,
    }

\usepackage[skip=2pt]{caption}
\begin{document}

\title{Single-View Shape Completion for Robotic Grasping in Clutter}
%
%
\author{Abhishek Kashyap\inst{1}\orcidID{0009-0004-2316-8831} \and
Yuxuan Yang\inst{1}\orcidID{0000-0003-1528-4301} \and
Henrik Andreasson\inst{1}\orcidID{0000-0002-2953-1564} \and
Todor Stoyanov\inst{1}\orcidID{0000-0002-6013-4874}}
\authorrunning{A. Kashyap et al.}
%
\institute{Örebro University, Örebro 70182, Sweden\\
\email{\{abhishek.kashyap,yuxuan.yang,henrik.andreasson,todor.stoyanov\}@oru.se}}
\maketitle              

\begin{abstract}


In vision-based robot manipulation, a single camera view can only capture one side of objects of interest, with additional occlusions in cluttered scenes further restricting visibility. As a result, the observed geometry is incomplete, and grasp estimation algorithms perform suboptimally. To address this limitation, we leverage diffusion models to perform category-level 3D shape completion from partial depth observations obtained from a single view, reconstructing complete object geometries to provide richer context for grasp planning. Our method focuses on common household items with diverse geometries, generating full 3D shapes that serve as input to downstream grasp inference networks. Unlike prior work, which primarily considers isolated objects or minimal clutter, we evaluate shape completion and grasping in realistic clutter scenarios with household objects. In preliminary evaluations on a cluttered scene, our approach consistently results in better grasp success rates than a naive baseline without shape completion by 23\% and over a recent state of the art shape completion approach by 19\%. Our code is available at \url{https://amm.aass.oru.se/shape-completion-grasping/}. 


\end{abstract}

\section{Introduction}
\label{section_introduction}

\begin{figure}[htpb]
  \centering
  \includegraphics[width=\textwidth]{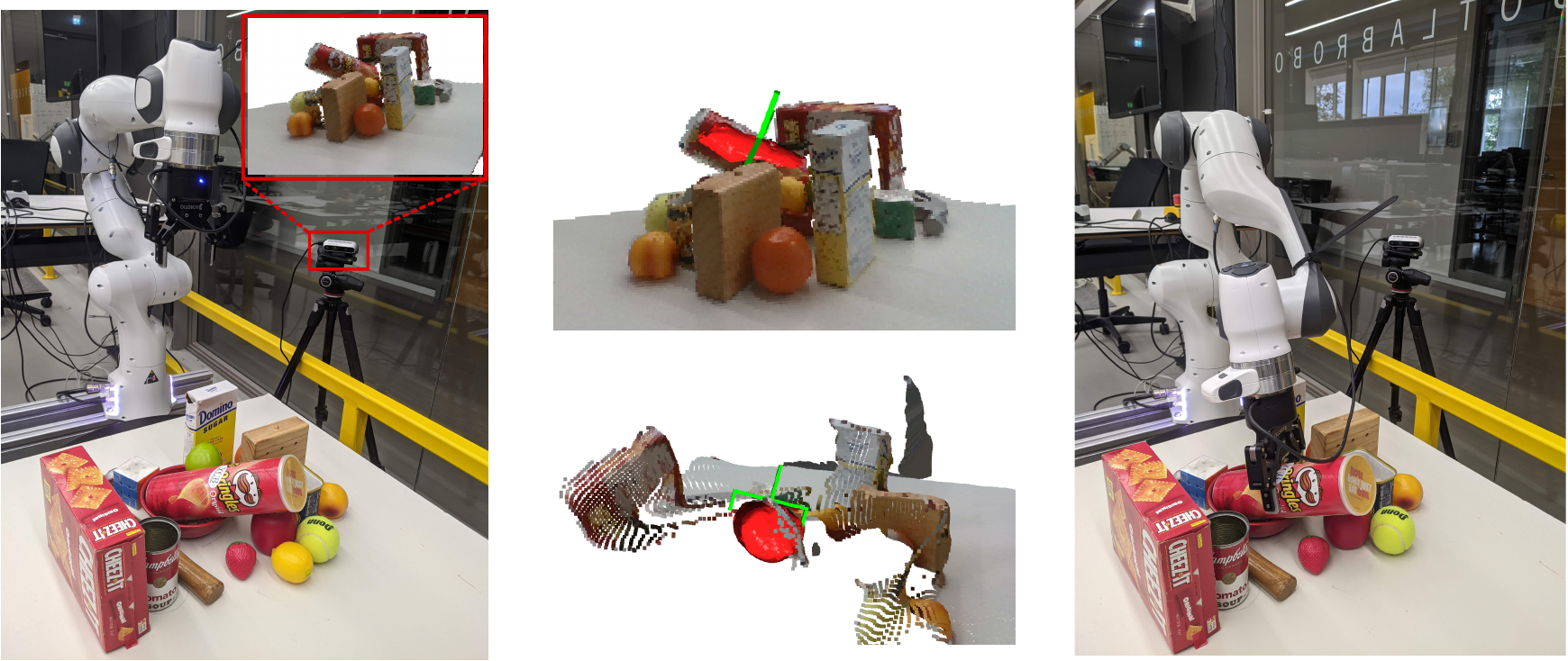}
  \caption{Grasping in clutter with shape completion. \textit{Left}: Household objects in robot workspace, viewed through Intel Realsense D435i. \textit{Middle}: Shape completion of the target object and grasp inference on the completed shape. \textit{Right}: Grasp execution.}
  \label{fig:opening}
\end{figure}

Autonomous grasping is the basis of many robot manipulation systems and has attracted substantial research in recent years \cite{newbury2023deep}. Despite significant progress, current state-of-the-art methods perform poorly in highly cluttered environments~\cite{zheng2024robocas}, such as those common in e.g., household robotics settings (see Fig.~\ref{fig:opening}). Because a single camera view captures only part of an object and clutter introduces further occlusions, surface visibility is limited, resulting in incomplete observations. Generating grasps with only partial geometry is prone to errors---resulting in grasps that are in collision or reliant on non-existent surfaces. Consequently, grasp generation from partial observations is often unreliable, highlighting the need for approaches that reason over complete object geometry.

In this paper, we address the problem of grasp generation for cluttered scenarios by leveraging surface generative models based on diffusion models~\cite{ho2020denoising}. By drawing on knowledge of previously observed objects, such generative models can estimate complete object shapes by filling in sections of the objects not visible to a camera, thus providing a complete geometric context for reliable grasp planning. We present a systems-level approach (see Fig.~\ref{fig:pipeline}) to the problem, where we combine multiple components to address grasping in clutter: acquiring scene information with an RGB-D camera, segmenting objects to be grasped, estimating their complete shapes, and inferring grasps on those complete shapes rather than on the partial observations captured by the camera.

While prior work has explored estimating object shapes from single-view RGB~\cite{wu2024unique3d,huang2025spar3d} or depth inputs~\cite{yan2022shapeformer}, these methods often assume that partial point clouds are already aligned in a canonical frame---a requirement that is impractical in real-world manipulation, where objects are encountered in arbitrary poses and under occlusion. This misalignment limits the effectiveness of shape completion models trained solely on canonical data to the real-world robot grasping task. Additionally, recent efforts to apply shape completion to robot manipulation~\cite{mohammadi20233dsgrasp,3dscarp,agrawal2023real,iwase2025zerograsp} have primarily focused on simplified settings, with limited exploration of cluttered, occluded environments or validation through real-robot experiments.

We address these challenges by leveraging diffusion-based generative models to reconstruct complete 3D shapes from partial, unaligned observations in realistic scenarios. Our models are trained to be robust to clutter and occlusion, resulting in improved shape completion and grasp planning in real-world robot manipulation settings. Our main contributions are:

\begin{itemize}
    \item We present a system-level integration for object grasping in cluttered scenes that combines three learning-based components: open-vocabulary object segmentation, diffusion-based shape completion from arbitrarily oriented partial point clouds, and a modular grasp generation module.
    \item We demonstrate the effectiveness of our approach in real robot experiments using actual camera observations, demonstrating that incorporating shape completion as a preprocessing step improves grasp success rates on diverse household objects in cluttered environments.
    \item We introduce the first integration of diffusion-based shape completion in robotic manipulation and devise training routines to make the method more robust to occlusions.
\end{itemize}

\section{Related Work}
\label{section_related_work}

Grasp planning in cluttered environments is challenging due to occlusions and incomplete object geometry. Prior work has shown that in such settings it is beneficial to use data from multiple viewpoints~\cite{zeng2017multi}, however, acquiring it may be constrained by workspace or time constraints. In contrast, single-view grasp generation is based on inherently incomplete geometric information, as portions of objects are occluded by clutter or viewpoint.

To address challenges in single-view grasp estimation, S4G directly regresses 6-DoF grasps from a single depth view using per-point scoring and pose regression~\cite{qin2020s4g}, which is computationally intense. GraspNet-1Billion addresses occlusions by training a network to generate feasible approach vectors~\cite{fang2020graspnet}, which is challenging  when the optimal grasp vector collides with occluded surfaces.

To deal with partial geometric information available from a single view, recent works like 3DSGrasp~\cite{mohammadi20233dsgrasp}, SCARP~\cite{3dscarp}, SceneGrasp~\cite{agrawal2023real}, and ZeroGrasp~\cite{iwase2025zerograsp} perform shape completion prior to grasp prediction. 3DSGrasp~\cite{mohammadi20233dsgrasp} is evaluated only in clutter-free settings on $10$ YCB dataset objects~\cite{calli2015ycb}, where occlusion is not a limiting factor. SCARP~\cite{3dscarp} evaluates grasps on 5 tabletop objects from the ShapeNet dataset~\cite{shapenet2015} but restricts testing to isolated objects in simulation, without addressing real-world depth sensor noise or occlusion challenges. SceneGrasp\cite{agrawal2023real} performs simultaneous shape reconstruction, pose estimation, and grasp prediction on the NOCS dataset~\cite{wang2019nocsdataset} containing 6 object categories, but evaluation scenarios involve minimal occlusion and lack a reliable grasp validation procedure. A recent work, ZeroGrasp~\cite{iwase2025zerograsp}, does both shape completion and grasp prediction, and performs evaluations in real-world settings. We evaluate and compare against ZeroGrasp and note that performance with a noisy depth sensor degrades compared to reported result.

The strength of our approach lies in category-level shape completion that performs under varying occlusion levels, coupled with comprehensive real-robot grasping validation. Unlike prior work that focuses on isolated objects or minimal clutter, we evaluate our complete pipeline from RGB-D input to grasp execution in realistic household clutter scenarios, demonstrating the practical benefits of complete shape information for robotic manipulation.

\section{Method}
\label{section_method}
We adopt a modular approach to grasping in clutter, decomposing the problem into scene acquisition, object segmentation, shape completion, and grasp inference. Shape completion models are typically trained on datasets of single objects rather than entire scenes. By segmenting the scene and completing each object individually, we leverage these models in a way that aligns with their training distribution, producing plausible object geometries for grasp planning. This modular pipeline also allows flexibility to swap different segmentation or grasping components without retraining the entire system, which would be difficult in a monolithic approach.

\begin{figure}[htpb]
\centering
\includegraphics[width=\textwidth]{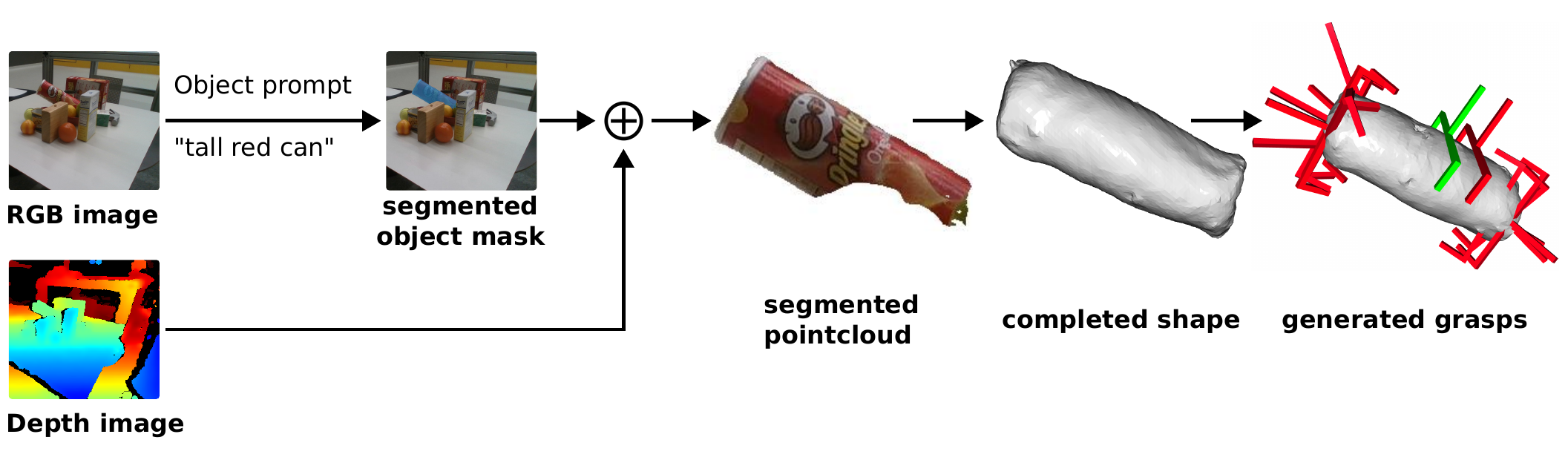}
\caption{Overview of the proposed method. RGB information is used to segment an object of interest. The object pointcloud is then fed into a diffusion model to obtain a completed surface, which then informs grasp planning. Grasps are ranked and selected for execution (green grasp in figure).}
\label{fig:pipeline}
\end{figure}

Given a robot manipulator, a set of household objects in its workspace, and a statically mounted RGB-D camera, target objects intended for grasping are specified and segmented via language prompts. Our complete pipeline, illustrated in Fig.~\ref{fig:pipeline}, operates as follows: the RGB component of the input RGB-D data undergoes language-guided segmentation to generate an object mask, leveraging contextual understanding to avoid over-segmentation and ensure complete object surfaces are identified. The mask is applied to the corresponding depth image to extract the visible point cloud of the target object. This partial point cloud is subsequently processed by a shape completion module that estimates the complete 3D geometry. The completed object shape serves as input to a grasp inference network, which predicts candidate grasps. Finally, a selected grasp is executed using a standard motion planner.

\subsection{Object segmentation}
\label{subsection_segmentation}

Object segmentation isolates an individual object's point cloud from the scene's point cloud and serves as the input to our shape completion model. Foundation models such as SAM2~\cite{ravi2024sam} have demonstrated remarkable performance in segmenting RGB images. However, they are still susceptible to over-segmentation and, in some cases, merging adjacent objects with similar visual properties. To address this, we employ LangSAM\footnote{https://github.com/luca-medeiros/lang-segment-anything}, which enables instance segmentation guided by short text prompts (e.g., “\textit{red bowl}”, “\textit{wooden block}”). This approach proved highly effective in producing precise masks of objects selected for grasping that were then used for extracting object point clouds from the scene.

\subsection{Single-view object reconstruction}
\label{subsection_recon_mesh}
While a point cloud captured from a single viewpoint provides only partial geometric information, it contains sufficient structural cues for a diffusion model trained on similar objects to infer the complete 3D shape, including regions occluded from the camera's view.

\subsubsection{Model architecture}
\label{subsubsection_model_arch}
We represent 3D objects using signed distance fields (SDFs), where each point in space is assigned its distance to the nearest surface, with the sign indicating whether the point lies inside or outside the object. This implicit surface representation is well-suited for learning-based reconstruction~\cite{park2019deepsdf,chou2022gensdf}.

We employ Diffusion-SDF~\cite{chou2023diffusion} to estimate complete object shapes from partial point clouds obtained through single-view depth sensing. The architecture consists of three core components: GenSDF~\cite{chou2022gensdf} for learning generalizable signed distance fields, a variational autoencoder (VAE)~\cite{kingma2014autoencoding} that compresses object shapes into compact latent representations, and a diffusion network that directly predicts denoised latent vectors~\cite{ramesh2022hierarchical}.

\subsubsection{Category-level shape completion}
\label{subsubsection_category_level_completion}
We find that category-level shape completion is essential to resolve fundamental ambiguities that arise when objects from different categories share similar local geometric features. For instance, a curved surface patch could plausibly belong to a bottle, mug, or bowl, while a flat surface segment might indicate either a box face or the side of a bottle with flat edges. Without category-level constraints, we observe that the shape completion process lacks sufficient context to disambiguate between these geometrically similar but distinct object types. We implement category-level completion by training an ensemble model, with a separate checkpoint for every category of objects (enumerated in Table~\ref{table_dataset_objs}). This strategy enables reliable shape inference and facilitates extensibility to new object categories through additional models in the ensemble.

\subsection{Grasp pose estimation}
\label{subsection_grasp_est}
A grasp pose is a 6-DOF end-effector pose for grasping an object. Given a point cloud input, grasp pose estimation predicts candidate grasps that guide the manipulator's end-effector to execute successful object grasps. Our pipeline's modular design enables integration with any point cloud-based grasp estimation method. We use the state-of-the-art diffusion-based GraspGen~\cite{murali2025graspgen} for predicting grasps on the completed object shapes. All predicted grasps are associated with a predicted grasp score.

Predicted grasps are divided into two categories based on their approach vectors: those that fall inside a $40\degree$ cone relative to the vertical, and those that fall outside. Grasps in both categories are ranked separately by their predicted scores. We select the top $K=5$ grasps with a preference for those falling within the cone, as these vertical approaches minimize collision risk with neighbouring objects and the base surface. If fewer than $K$ grasps are inside the cone, additional grasps from outside the cone are selected to reach a total of $K$. We then cycle through these top $K$ grasps and pass them to the motion planner, terminating when the motion planner succeeds. This multi-attempt strategy improves robustness by handling motion planning failures that arise from practical constraints such as arm reachability and layout of objects in the workspace.

\section{Experiments and Results}
\label{section_experiments_results}

\subsection{Implementation Details}
\label{subsection_implementation}

\subsubsection{Dataset}
\label{subsubsection_dataset}
Although large-scale 3D object datasets such as Objaverse~\cite{deitke2023objaverse} contain diverse object categories, we found that household categories relevant to robotic grasping lacked sufficient samples for effective training. To address this, we constructed our dataset by selecting subsets of objects from three synthetic model collections: 3DNet~\cite{wohlkinger20123dnet}, ShapeNetCore~\cite{shapenet2015}, and HouseCat6D~\cite{jung2024housecat6d}.

For evaluation, we focused on six categories with adequate training samples to ensure reliable shape completion performance: \textbf{apple, bottle, bowl, box, can, and hammer} (Table~\ref{table_dataset_objs}). We manually excluded samples that were semantically inconsistent with their designated category or that produced invalid or artifact-prone meshes when reconstructed by extracting iso-surfaces from their signed distance fields.

Our data preprocessing pipeline followed several steps to ensure mesh quality and realistic training conditions. Many of the original meshes were not consistently watertight and contained small gaps or holes in the surface. Such defects prevent a clear distinction between the interior and exterior of the object, which is essential when computing SDFs. Similarly, surface normals were often oriented inconsistently, pointing inward on some faces and outward on others. Correct normal orientation ensures a coherent surface description and avoids errors in downstream geometry processing. We used mesh2sdf\footnote{https://github.com/wang-ps/mesh2sdf} for watertight conversion and trimesh\footnote{https://trimesh.org/} to correct surface normal orientations. To generate realistic partial point clouds for training, we then applied random rotations to each mesh and performed virtual camera raycasting using Open3D\footnote{https://www.open3d.org/} to simulate real-world depth sensing conditions. This raycasting approach better captures occlusions and line-of-sight visibility constraints compared to alternative methods such as distance-based point filtering or depth-sorted point selection.

\begin{table}[h]
\caption{Sample counts for selected household objects by dataset.}
\label{table_dataset_objs}
\begin{center}
\begin{tabular}{c|c|c|c}
Object Category & 3DNet~\cite{wohlkinger20123dnet} & ShapeNetCore~\cite{shapenet2015} & HouseCat6D~\cite{jung2024housecat6d} \\
\hline
Apple & 12 & - & - \\
Bottle & 74 & 498 & 21 \\
Bowl & 31 & 186 & - \\
Box & - & - & 23 \\
Can & - & 108 & 23 \\
Hammer & 36 & - & - \\
\hline
\end{tabular}
\end{center}
\end{table}

\subsubsection{Training details}
\label{subsubsection_training_details}
We followed the original 3-stage training procedure of Diffusion-SDF~\cite{chou2023diffusion}. Training the network took approximately three days per object category on an NVIDIA A40 GPU with 48 GB of VRAM. We use LangSAM and GraspGen~\cite{murali2025graspgen} with their provided model weights. As our focus is not on developing novel segmentation or grasp prediction, we employ pre-trained models trained on large-scale datasets with proven generalization. In contrast, Diffusion-SDF required retraining on our specific dataset to accurately reconstruct shapes of the target object categories and handle instances in arbitrary orientations, as described in section~\ref{subsubsection_category_level_completion}.

\subsection{Results}
\label{subsection_results}

To evaluate our approach, we first validate the object reconstruction module on an existing data set and then proceed with real-world robot experiments for evaluating grasp success rates of the full system.

\subsubsection{Reconstruction Quality}
\label{subsubsection_eval_datasets}

We validate our 3D reconstruction capabilities on the ReOcS real-world dataset~\cite{iwase2025zerograsp} containing household items spread out in various configurations across three difficulty levels based on clutter and occlusion: \textit{easy}, \textit{normal}, and \textit{hard}. Following ZeroGrasp~\cite{iwase2024zero}, we use bidirectional Chamfer distance as our evaluation metric.

\begin{figure}[thpb]
    \centering

    \begin{minipage}{0.1\textwidth}
        \centering
        \textbf{Clutter}
    \end{minipage}
    \hfill
    \begin{minipage}{0.22\textwidth}
        \centering
        \textbf{RGB images}
    \end{minipage}
    \hfill
    \begin{minipage}{0.51\textwidth}
        \centering
        \textbf{Reconstructions}
    \end{minipage}

    \begin{minipage}{0.1\textwidth}
    \end{minipage} 
    \hfill
    \begin{minipage}{0.22\textwidth}
    \end{minipage} 
    \hfill
    \begin{minipage}{0.51\textwidth}
        \centering
        \begin{minipage}{0.48\textwidth}
            \centering
            (top view)
        \end{minipage}
        \hfill
        \begin{minipage}{0.48\textwidth}
            \centering
            (bottom view)
        \end{minipage}
    \end{minipage}

    \noindent\rule{\textwidth}{0.4pt}

    \begin{minipage}{0.1\textwidth}
        \centering
        Easy
    \end{minipage}
    \hfill
    \begin{minipage}{0.22\textwidth}
        \centering
        \includegraphics[width=\textwidth]{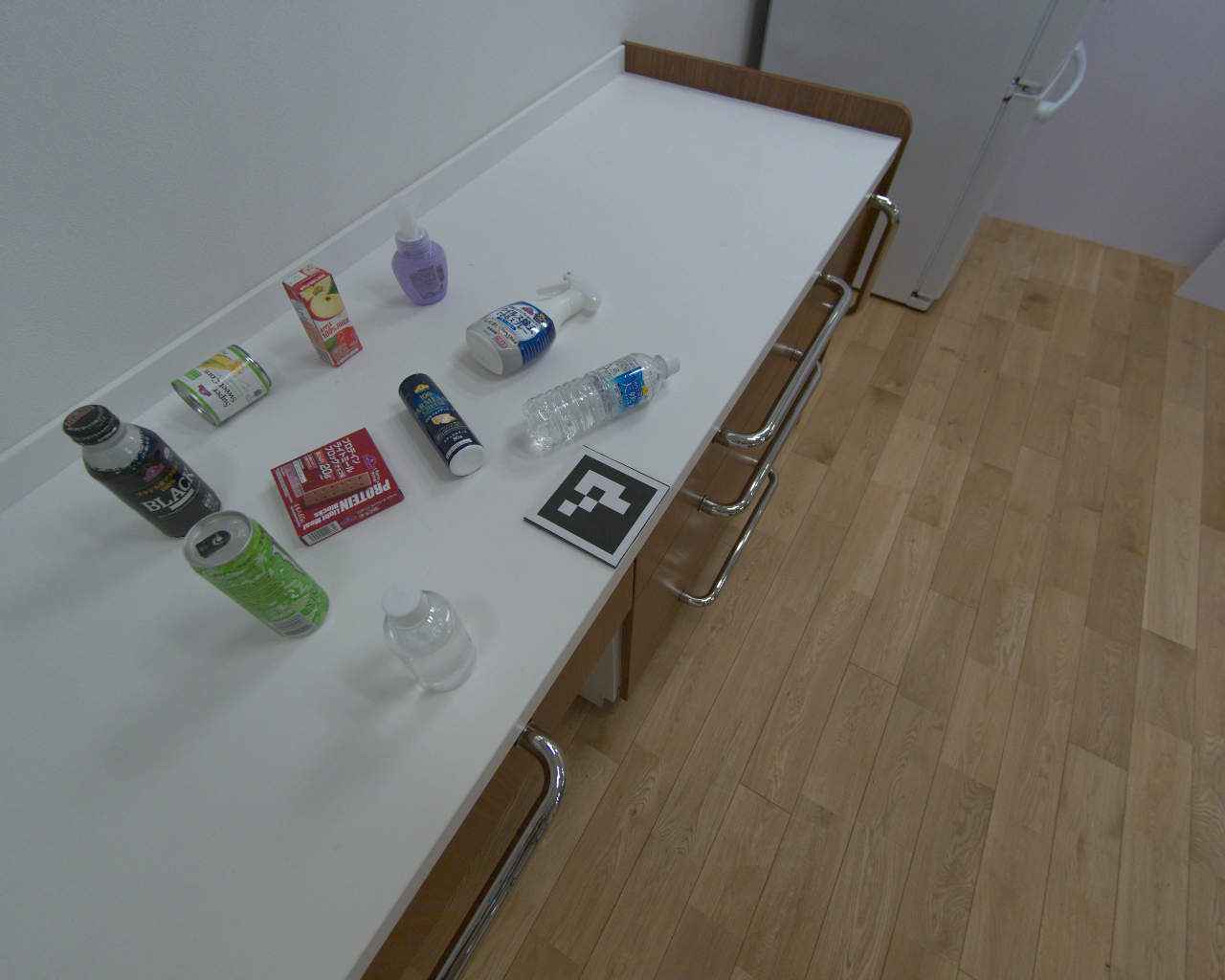}
    \end{minipage}
    \hfill
    \begin{minipage}{0.22\textwidth}
        \centering
        \includegraphics[width=\textwidth]{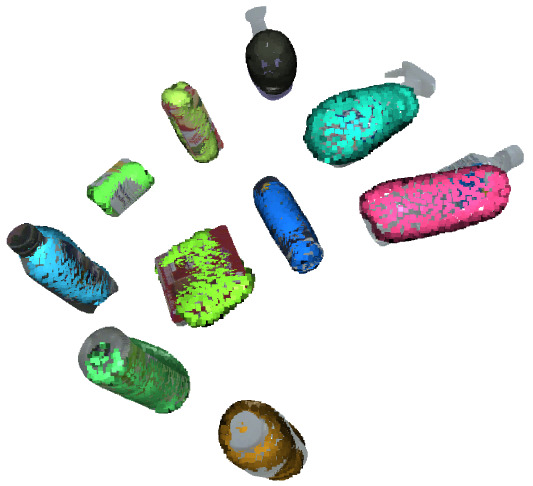}
    \end{minipage}
    \hfill
    \begin{minipage}{0.22\textwidth}
        \centering
        \includegraphics[width=\textwidth]{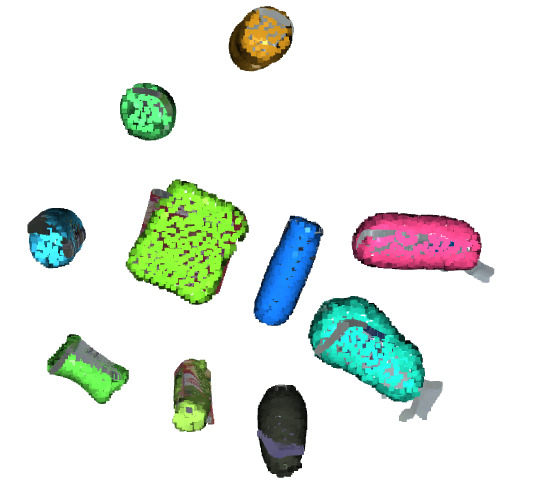}
    \end{minipage}

    \noindent\hdashrule{\textwidth}{0.4pt}{2pt}

    \begin{minipage}{0.1\textwidth}
        \centering
        Normal
    \end{minipage}
    \hfill
    \begin{minipage}{0.22\textwidth}
        \centering
        \includegraphics[width=\textwidth]{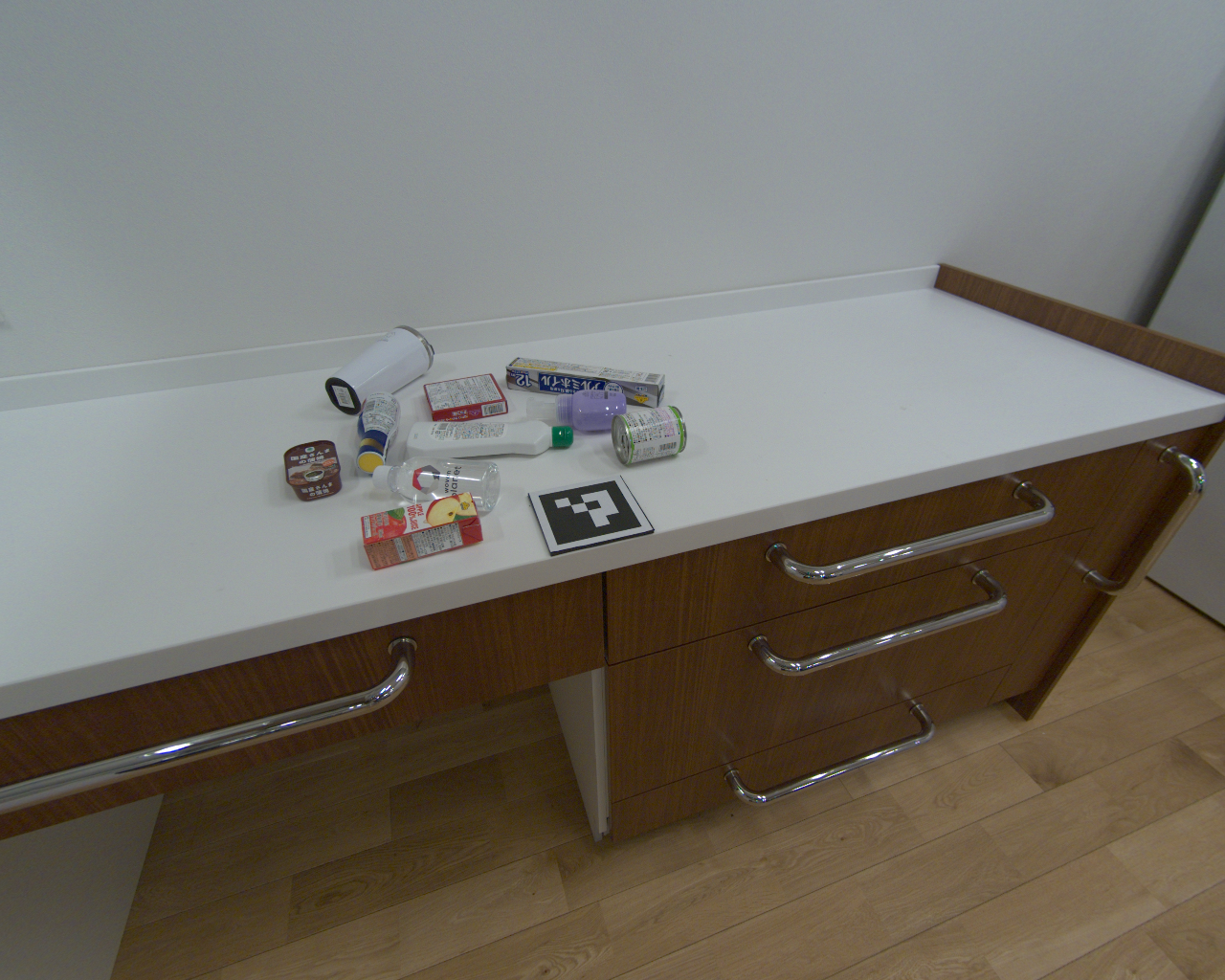}
    \end{minipage}
    \hfill
    \begin{minipage}{0.22\textwidth}
        \centering
        \includegraphics[width=\textwidth]{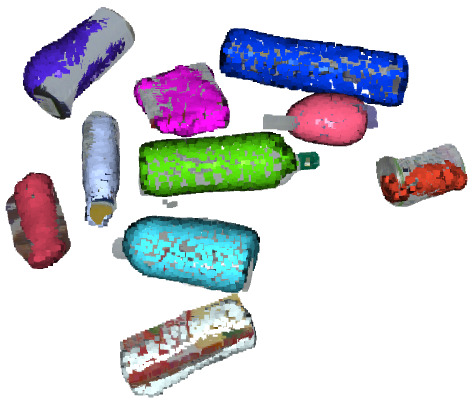}
    \end{minipage}
    \hfill
    \begin{minipage}{0.22\textwidth}
        \centering
        \includegraphics[width=\textwidth]{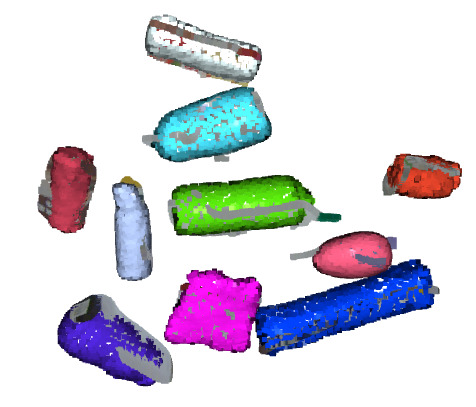}
    \end{minipage}

    \noindent\hdashrule{\textwidth}{0.4pt}{2pt}

    \begin{minipage}{0.1\textwidth}
        \centering
        Hard
    \end{minipage}
    \hfill
    \begin{minipage}{0.22\textwidth}
        \centering
        \includegraphics[width=\textwidth]{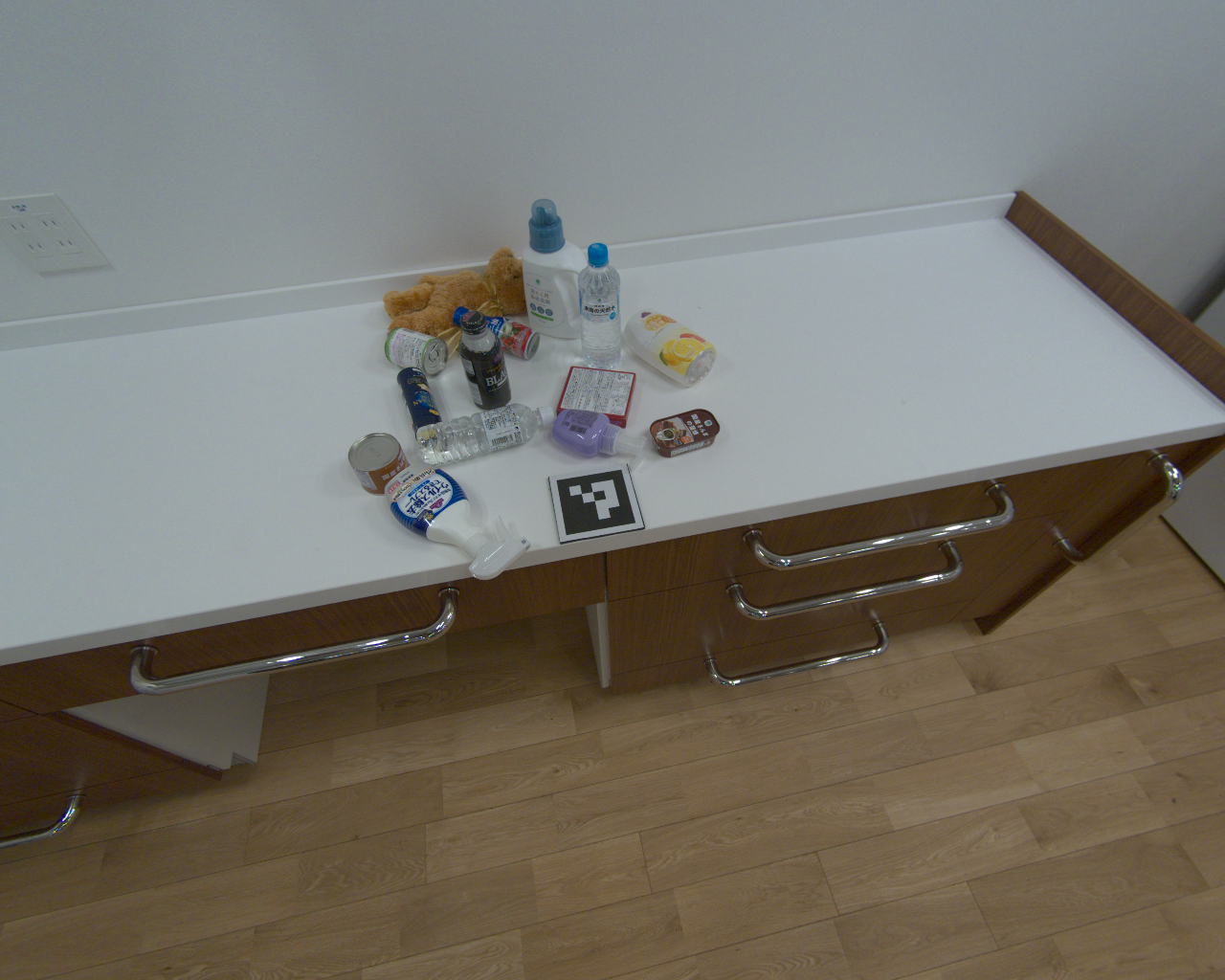}
    \end{minipage}
    \hfill
    \begin{minipage}{0.22\textwidth}
        \centering
        \includegraphics[width=\textwidth]{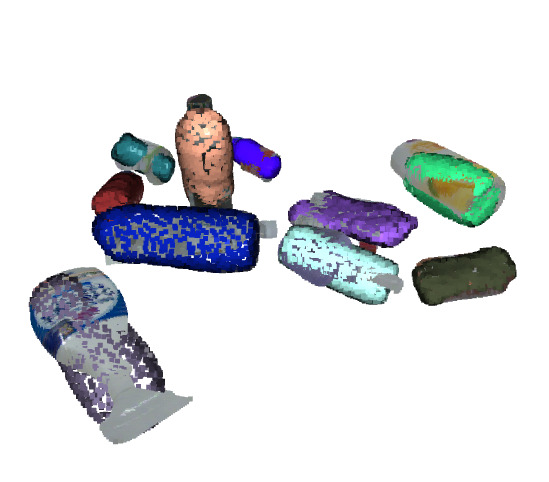}
    \end{minipage}
    \hfill
    \begin{minipage}{0.22\textwidth}
        \centering
        \includegraphics[width=\textwidth]{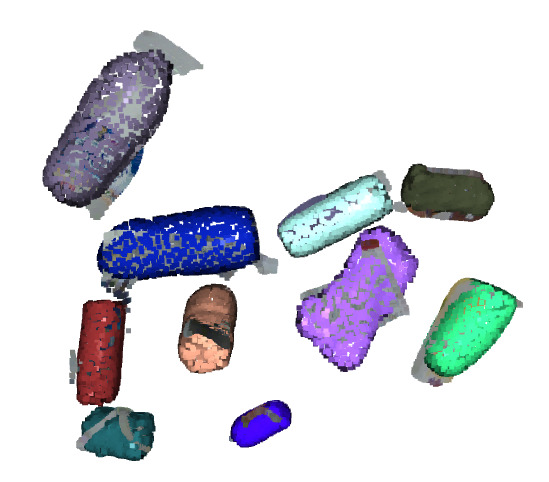}
    \end{minipage}
    
    \noindent\rule{\textwidth}{0.4pt}
    
    \caption{Qualitative results of Diffusion-SDF on different levels of clutter (easy, normal, and hard) of the ReOcS dataset~\cite{iwase2025zerograsp}.}
    \label{fig:eval_qual}
\end{figure}

Table~\ref{table_chamfer_comparison} presents Chamfer distances and reconstruction success rates for objects in the ReOcS dataset that belong to the categories used in our shape completion training (see Table~\ref{table_dataset_objs}). We define a reconstruction as successful if it yields a valid mesh, which requires sufficient output points for ZeroGrasp and a signed distance field from which a mesh can be extracted for Diffusion-SDF. Although ZeroGrasp achieved lower Chamfer distances, their released checkpoint failed for approximately 30-35\% of samples for unknown reasons whereas our model reconstructed all instances. Presented Chamfer distances are for those instances that were successfully reconstructed by both ZeroGrasp and our model. Additionally, qualitative results in Fig.~\ref{fig:eval_qual} demonstrate that our approach produces reasonable surface completions, enabling us to proceed with real-world grasping evaluations.

\begin{table}[thpb]
\caption{Comparison of reconstruction quality and success rates across object types and difficulty levels on the ReOcS dataset.}
\centering
\setlength{\tabcolsep}{5pt}
\begin{tabular}{c|ccc:ccc|cc}
\hline
\multirow{3}{*}{Clutter} & \multicolumn{6}{c|}{Chamfer distance (in mm) $\downarrow$} & \multicolumn{2}{c}{\multirow{2}{*}{Reconstruction Success \%}} \\
\cline{2-7}
& \multicolumn{3}{c:}{ZeroGrasp} & \multicolumn{3}{c|}{Ours} & & \\
& bottle & box & can & bottle & box & can & ZeroGrasp & Ours \\
\hline
Easy & 10.44 & 8.92 & 7.92 & 16.75 & 11.49 & 14.90 & 62.34 & 100 \\
Normal & 8.75 & 8.69 & 8.74 & 15.95 & 12.71 & 15.56 & 64.77 & 100 \\
Hard & 10.12 & 10.21 & 9.28 & 16.90 & 15.23 & 17.53 & 69.85 & 100 \\
\hline
\end{tabular}
\label{table_chamfer_comparison}
\end{table}

\begin{figure}[thpb]
    \centering
    \begin{subfigure}{0.45\textwidth}
        \centering
        \includegraphics[width=\textwidth]{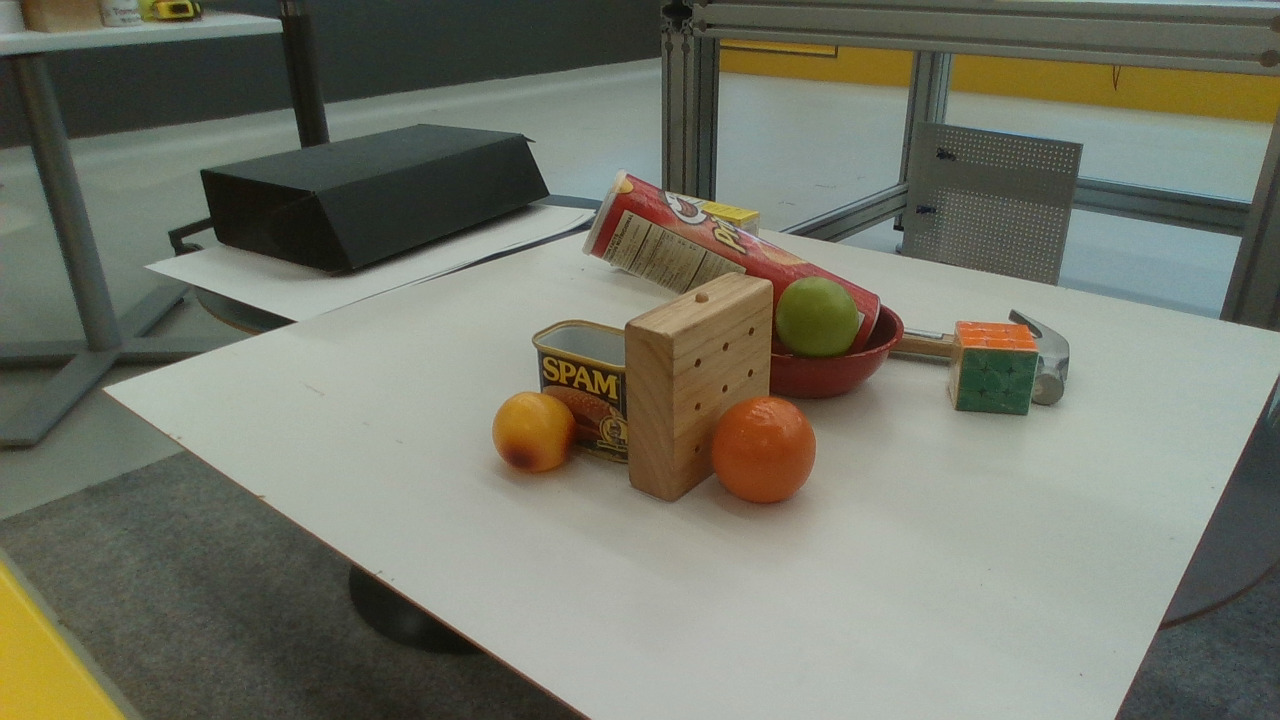}
        \caption{Target objects: pringles can, wooden block}
        \label{fig:sub1}
    \end{subfigure}
    \hfill
    \begin{subfigure}{0.45\textwidth}
        \centering
        \includegraphics[width=\textwidth]{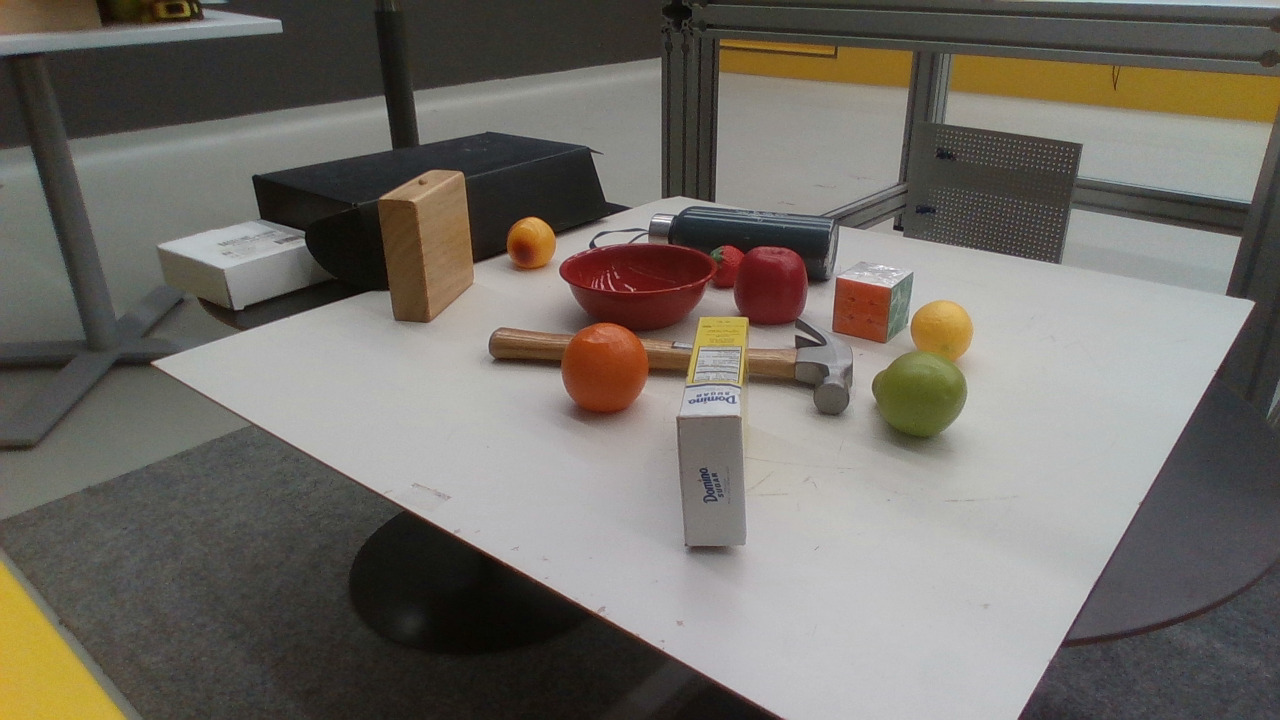}
        \caption{Target objects: hammer, bowl, apple, bottle}
        \label{fig:sub2}
    \end{subfigure}
    \caption{Scene configurations used in the real robot experiments.}
    \label{fig:real_robot_setups}
\end{figure}

\subsubsection{Grasping in Clutter}
\label{subsubsection_eval_real_robot}
We evaluate the full pipeline through grasping experiments on a Franka Emika Panda robot equipped with a Robotiq 2F-85 gripper. We use ROS2 (Humble)~\cite{ros2} as a middleware and plan robot motions using MoveIt2~\cite{coleman2014moveit}. We evaluate on two different experimental setups, as shown in Fig.~\ref{fig:real_robot_setups}, designed to achieve appropriate occlusion levels across all object categories. A single setting would either under-occlude simpler geometries or over-occlude challenging shapes like hammers and bowls, preventing fair comparison across categories. Compared to the ReOcS dataset's \textit{easy}, \textit{normal}, and \textit{hard} splits~\cite{iwase2025zerograsp}, both our layouts would qualify as normal-hard.

We evaluate the quality of the proposed grasps by executing the highest-ranked grasp and recording the percentage of successful grasps, with $10$ trials per object category in the scene. We judge grasps to be successful if the target object remains grasped after post-grasp lift-up for longer than 5 seconds. Quantitative results are presented in Table \ref{table_real_robot_exp}.

\begin{figure}[thpb]
\newcommand{\imgheight}{1cm} 
\centering
\vspace{-0.05 cm}
\begin{minipage}[t]{0.22\textwidth}
\centering
\textbf{Object} \\[0.5ex]
\vspace{0.1cm} 
"wooden block" \\ (box)
\end{minipage}%
\begin{minipage}[t]{0.25\textwidth}
\centering
\textbf{Input} \\[0.5ex]
\includegraphics[height=\imgheight]{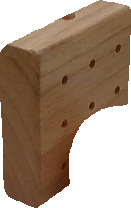}
\end{minipage}%
\begin{minipage}[t]{0.25\textwidth}
\centering
\textbf{ZeroGrasp}~\cite{iwase2025zerograsp} \\[0.5ex]
\includegraphics[height=\imgheight]{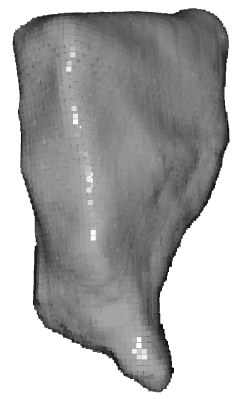}
\end{minipage}%
\begin{minipage}[t]{0.25\textwidth}
\centering
\textbf{Ours} \\[0.5ex]
\includegraphics[height=\imgheight]{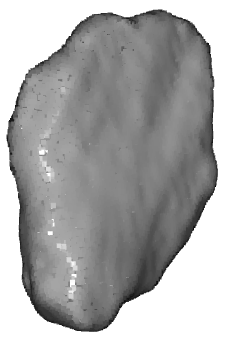}
\end{minipage}

\vspace{2ex} 

\begin{minipage}[t]{0.22\textwidth}
\centering
\vspace{-0.8cm}
"pringles can" \\ (can)
\end{minipage}%
\begin{minipage}[t]{0.25\textwidth}
\centering
\includegraphics[height=\imgheight]{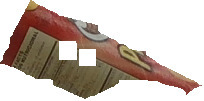}
\end{minipage}%
\begin{minipage}[t]{0.25\textwidth}
\centering
\includegraphics[height=\imgheight]{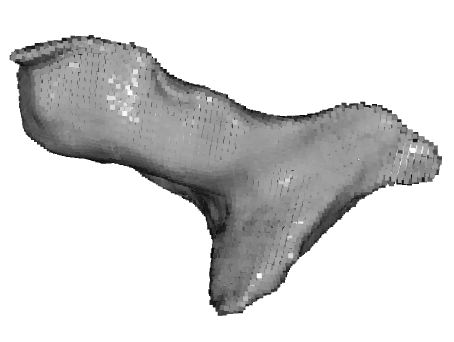}
\end{minipage}%
\begin{minipage}[t]{0.25\textwidth}
\centering
\includegraphics[height=\imgheight]{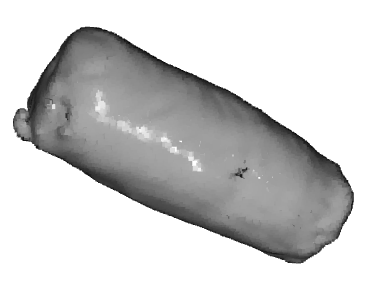}
\end{minipage}

\vspace{2ex} 

\begin{minipage}[t]{0.22\textwidth}
\centering
\vspace{-0.8cm}
"red apple" \\ (apple)
\end{minipage}%
\begin{minipage}[t]{0.25\textwidth}
\centering
\includegraphics[height=\imgheight]{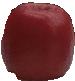}
\end{minipage}%
\begin{minipage}[t]{0.25\textwidth}
\centering
\includegraphics[height=\imgheight]{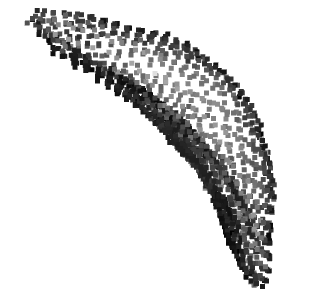}
\end{minipage}%
\begin{minipage}[t]{0.25\textwidth}
\centering
\includegraphics[height=\imgheight]{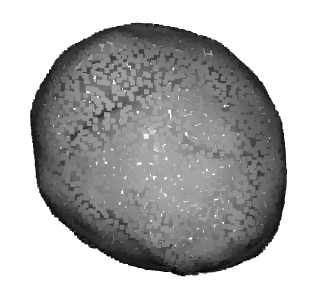}
\end{minipage}

\vspace{2ex} 

\begin{minipage}[t]{0.22\textwidth}
\centering
\vspace{-0.8cm}
"navy blue \\ bottle" \\ (bottle)
\end{minipage}%
\begin{minipage}[t]{0.25\textwidth}
\centering
\includegraphics[width=0.9\textwidth]{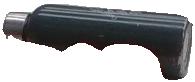}
\end{minipage}%
\begin{minipage}[t]{0.25\textwidth}
\centering
\includegraphics[height=\imgheight]{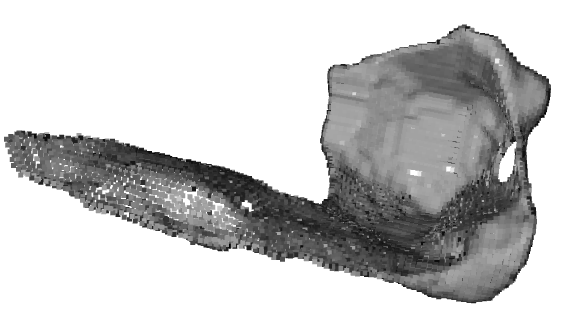}
\end{minipage}%
\begin{minipage}[t]{0.25\textwidth}
\centering
\includegraphics[height=\imgheight]{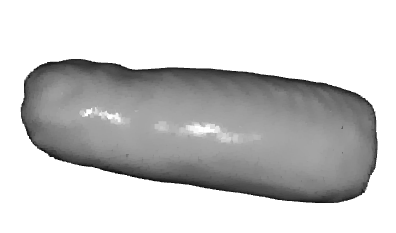}
\end{minipage}

\vspace{2ex} 

\begin{minipage}[t]{0.22\textwidth}
\centering
\vspace{-0.8cm}
"red bowl" \\ (bowl)
\end{minipage}%
\begin{minipage}[t]{0.25\textwidth}
\centering
\includegraphics[height=\imgheight]{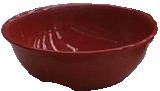}
\end{minipage}%
\begin{minipage}[t]{0.25\textwidth}
\centering
\includegraphics[height=\imgheight]{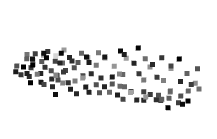}
\end{minipage}%
\begin{minipage}[t]{0.25\textwidth}
\centering
\includegraphics[height=\imgheight]{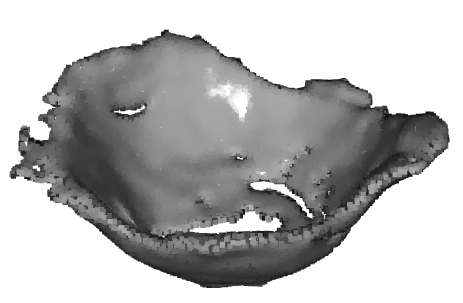}
\end{minipage}

\vspace{2ex} 

\begin{minipage}[t]{0.22\textwidth}
\centering
\vspace{-1cm}
"hammer" \\ (hammer)
\end{minipage}%
\begin{minipage}[t]{0.25\textwidth}
\centering
\includegraphics[width=0.8\textwidth]{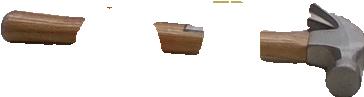}
\end{minipage}%
\begin{minipage}[t]{0.25\textwidth}
\centering
\includegraphics[height=\imgheight]{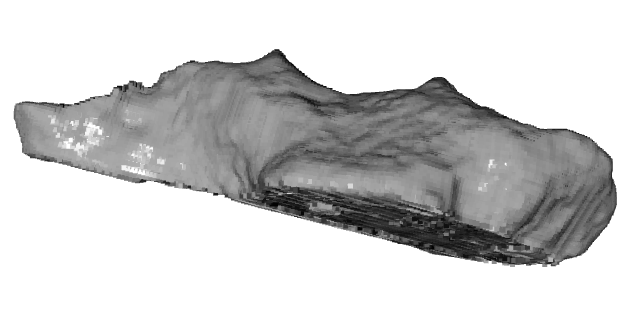}
\end{minipage}%
\begin{minipage}[t]{0.25\textwidth}
\centering
\includegraphics[height=\imgheight]{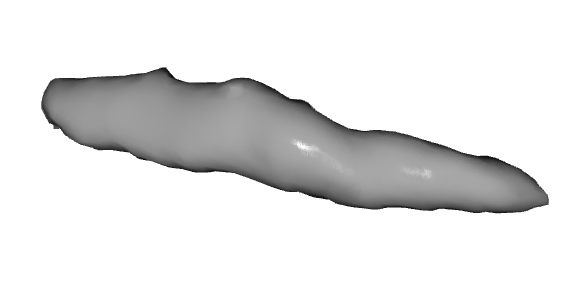}
\end{minipage}
\caption{Comparison of reconstruction quality from real-world experiments. Our approach consistently results in more plausible geometries.}
\label{fig:reconstruction_comparison}
\vspace{-0.05 cm}
\end{figure}

We compare against ZeroGrasp~\cite{iwase2025zerograsp} as our baseline, using their complete pipeline of reconstruction followed by grasp estimation. The proposed grasps are then ranked and selected in the same manner as those from our system, following the procedure in Section~\ref{subsection_grasp_est}. Qualitative comparison in Fig.~\ref{fig:reconstruction_comparison} reveals significant limitations in ZeroGrasp's reconstruction quality. Several reported successes, particularly for ``can'' and ``apple'' categories, likely stem from incidental factors rather than reliable reconstruction, as evidenced by the poor shape quality shown.

\begin{table}[thpb]
\caption{Grasp success rates with and without shape completion, compared to ZeroGrasp~\cite{iwase2025zerograsp}. Results shown as successful/failed grasps ($S$/$F$) and success rate ($S\%$) across $10$ trials per object category.}
\label{table_real_robot_exp}
\begin{center}
\small
\begin{tabularx}{\textwidth}{>{\centering\arraybackslash}p{0.15\textwidth}|*{3}{>{\centering\arraybackslash}X}|*{3}{>{\centering\arraybackslash}X}|*{3}{>{\centering\arraybackslash}X}}
\hline
\multirow{2}{*}{\begin{tabular}{c} Object \\ category \end{tabular}} 
& \multicolumn{3}{c|}{GraspGen} & \multicolumn{3}{c|}{Ours } & \multicolumn{3}{c}{ZeroGrasp} \\
& \multicolumn{3}{c|}{(no shape completion)} & \multicolumn{3}{c|}{(shape completion)} & \multicolumn{3}{c}{(shape completion)}\\
\cline{2-10}
& $S$ & $F$ & $S\%$ & $S$ & $F$ & $S\%$ & $S$ & $F$ & $S\%$ \\
\hline
apple & 6 & 4 & 60 & 10 & 0 & \textbf{100} & 8 & 2 & 80 \\
bottle & 7 & 3 & \textbf{70} & 7 & 3 & \textbf{70} & 2 & 8 & 20 \\
bowl & 6 & 4 & 60 & 9 & 1 & \textbf{90} & 5 & 5 & 50 \\
box & 7 & 3 & 70 & 10 & 0 & \textbf{100} & 10 & 0 & \textbf{100} \\
can & 5 & 5 & 50 & 6 & 4 & 60 & 7 & 3 & \textbf{70} \\
hammer & 3 & 7 & 30 & 6 & 4 & \textbf{60} & 5 & 5 & 50 \\
\hline
Average & & & 56.67 & & & \textbf{80} & & & 61.67 \\
\hline
\end{tabularx}
\end{center}
\end{table}

\subsection{Inference time}
\label{subsection_inference_time}
On the NVIDIA RTX 2000 Ada Generation Laptop GPU (8~GB VRAM), the full pipeline as depicted in Fig.~\ref{fig:pipeline}  requires approximately 4--5~s: object segmentation 0.8~s, shape completion 3~s, alignment 0.2~s, and grasp estimation 0.4–0.6~s. By comparison, ZeroGrasp achieves an inference time of 2--3~s; however, while faster, it does not yield reconstructions of comparable quality. According to their reported results, inference on an NVIDIA A100 achieves 212~ms, with GPU memory usage remaining below 8~GB.

\section{Conclusion}
\label{section_conclusion}
In this paper, we present a systems-level approach for improved object grasping in clutter by exploiting generative capabilities of diffusion models. We demonstrate through real-robot experiments that shape completion significantly improves grasping success across diverse household objects. Our diffusion-based approach reliably reconstructs complete geometries from partial observations, beating the baseline by 19\% and leading to measurably better grasp performance compared to methods without shape completion.

While our category-level approach requires separate models per object type, it enables straightforward extension to new categories. Key directions for future work include improving model generality through approaches like language-aligned models guiding shape completion and reducing inference times to enable real-time robotic applications.

\begin{credits}
\subsubsection{\ackname} This work was partially supported by the Wallenberg AI, Autonomous Systems and Software Program (WASP) funded by the Knut and Alice Wallenberg Foundation. The computations and data handling were enabled by resources provided by the National Academic Infrastructure for Supercomputing in Sweden (NAISS), partially funded by the Swedish Research Council through grant agreement no. 2022-06725.
\subsubsection{\discintname} The authors have no competing interests to declare that are relevant to the content of this article.
\end{credits}

\bibliographystyle{splncs04}
\bibliography{references}

\end{document}